\documentclass[a4paper,conference]{IEEEtran}
\ifCLASSINFOpdf
\else
\fi
\usepackage{xcolor}

\usepackage{graphicx}
\usepackage{amsmath}
\usepackage{amssymb}
\usepackage{algorithmic,algorithm}
\usepackage{xspace}

\ifCLASSOPTIONcompsoc
  \usepackage[caption=false,font=normalsize,labelfont=sf,textfont=sf]{subfig}
\else
  \usepackage[caption=false,font=footnotesize]{subfig}
\fi
\usepackage{cite}

\begin{document}
%
\title{Layer-Wise Data-Free CNN Compression}

\author{\IEEEauthorblockN{Maxwell Horton\\ Yanzi Jin\\ Ali Farhadi\\ and Mohammad Rastegari}
\IEEEauthorblockA{Apple \\
Email: mchorton@apple.com}
}

\maketitle

\newcommand{\bn}[1]{\mathcal{B}_{#1}}
\newcommand{\conv}[1]{\mathcal{C}_{#1}}

\begin{abstract}
We present a computationally efficient method for compressing a trained neural network without using real data. We break the problem of data-free network compression into independent layer-wise compressions. We show how to efficiently generate layer-wise training data using only a pretrained network. We use this data to perform independent layer-wise compressions on the pretrained network. We also show how to precondition the network to improve the accuracy of our layer-wise compression method. We present results for layer-wise compression using quantization and pruning. When quantizing, we compress with higher accuracy than related works while using orders of magnitude less compute. When compressing MobileNetV2 and evaluating on ImageNet, our method outperforms existing methods for quantization at all bit-widths, achieving a $+0.34\%$ improvement in $8$-bit quantization, and a stronger improvement at lower bit-widths (up to a $+28.50\%$ improvement at $5$ bits). When pruning, we outperform baselines of a similar compute envelope, achieving $1.5$ times the sparsity rate at the same accuracy. We also show how to combine our efficient method with high-compute generative methods to improve upon their results.
\end{abstract}

\IEEEpeerreviewmaketitle

\section{Introduction}
The increasing popularity of Convolutional Neural Networks (CNNs) for visual recognition has driven the development of efficient networks capable of running on low-compute devices \cite{MobileNet,yolov2,xnornet}. Domains such as smart-home security, factory automation, and mobile applications often require efficient networks capable of running on edge devices rather than running on expensive, high-latency cloud infrastructure. Quantization \cite{google_quantization,xnornet} and pruning \cite{STR,DNW,GMP,RIGL} are two of the main areas of active research focused on compressing CNNs for improved execution time and reduced memory usage.

Most methods for CNN compression require retraining on the original training set. Applying post-training quantization usually results in poor network accuracy \cite{google_quantization} unless special care is taken in adjusting network weights \cite{DFQ}. Low-bit quantization provides additional challenges when data is not available. Popular methods for compressing through sparsity \cite{STR,DNW,GMP,RIGL} also require training on the original data. However, many real-world scenarios require compression of an existing model but prohibit access to the original dataset. For example, data may be legally sensitive or may have privacy restrictions \cite{privacy}. Additionally, data may not be available if a model which has already been deployed needs to be compressed.

As CNNs move from cloud computing centers to edge devices, the need for efficient algorithms for compression has also arisen. The increasing popularity of federated learning \cite{federated} has emphasized the importance of efficient on-device machine learning algorithms. Additionally, models deployed to edge devices may need to be compressed on-the-fly to support certain use cases. For example, models running on edge devices in areas of low connectivity may need to compress themselves when battery power is low. Such edge devices usually do not have enough storage to hold a variety of models with different performance profiles \cite{edgehardware}. To enable these use cases requires efficient data-free network compression.

We propose a simple and efficient method for data-free network compression. Our method is a layer-wise optimization based on the teacher-student paradigm \cite{KnowledgeDistillation}. A pretrained model is used as a ``teacher'' that will help train a compressed ``student'' model. During our layer-wise optimization, we generate data to approximate the input to a layer of the teacher network. We use this data to optimize the corresponding layer in the compressed student network. Figures ~\ref{alg} and ~\ref{fig:layerwise_optim} illustrate an overview of our method.

Our contributions are as follows. (1) We demonstrate that network compression can be broken into a series of layer-wise compression problems. (2) We develop a generic, computationally efficient algorithm for compressing a network using generated data. (3) We demonstrate the efficacy of our algorithm when compressing with quantization and pruning. (4) We achieve higher accuracy in our compressed models than related works.

\section{Related Work}
\textbf{Pruning and Quantization:} Quantization involves reducing the numerical precision of weights and/or activations from 32-bit floating point to a lower-bit integral representation \cite{tf_whitepaper,HMQ,DTQ,DSQ,FQN,integer,bhandare,google_quantization,tf_whitepaper,yang,xnornet,bnn,Gugli,tbnn} to reduce size and improve execution time. The most commonly used quantization scheme is affine quantization \cite{google_quantization,tf_whitepaper}. Pruning involves the deletion of weights from a neural network to improve efficiency. Many formulations exist \cite{RIGL,SNIP,OBS,DM,OBD,FilterSketch}. See \cite{STR} for a comprehensive overview.

\textbf{Data-Light and Data-Free Compression:} A few recent works have explored methods for quantizing a model using little data (``data-light'' methods) or no data (``data-free'' methods). In the data-light method AdaRound \cite{AdaRound}, the authors devise a method for optimizing the rounding choices made when quantizing a weight matrix. In the data-free method Data-Free Quantization \cite{DFQ}, the authors manipulate network weights and biases to reduce the post-training quantization error. Other equalization formulations have been explored \cite{WeightFactorization}.

A few works have explored data-light and data-free pruning. In the data-light method PFA \cite{PFA}, activation correlations are used to prune filters. An iterative data-free pruning method is presented in \cite{DFPP}, though they only prune fully-connected layers. Overparameterized networks are explored in \cite{OBD} and \cite{OBS}. Layer-wise approaches are explored in \cite{LOBS,wootz}.

\textbf{Generative Methods for Data-Free Compression:} Given a trained model, it's possible to create synthetic images that match characteristics of the training set's statistics. These images can be used to retrain a more efficient model.

In the data-free method Deep Inversion (DI) \cite{DI}, the authors use a pretrained model to generate a dataset which is then used to train a sparse model. A set of noise images are trained to match the network's BatchNorm \cite{BatchNorm} statistics, and to look realistic. The authors use these images to train a sparse model using Knowledge Distillation \cite{KnowledgeDistillation}. A similar data-light method appears in The Knowledge Within \cite{KnowledgeWithin}.

In Adversarial Knowledge Distillation \cite{AKD}, the authors generate synthetic images for model training using a GAN \cite{GAN}. The images are used in conjunction with Knowledge Distillation \cite{KnowledgeDistillation} to train a network. Similar GAN-based formulations have been developed \cite{GLBDFQ,dfnp,gzsq,dsg,zskt}.

In ZeroQ \cite{zeroq}, the authors tune noise images to match BatchNorm statistics of a network. Those images are then used to set quantizers. This method runs quickly, but requires substantial memory for backpropagation (Figure~\ref{fig:runtime_comparison}).

\begin{figure}[t]
\hrule
\vspace{0.1cm}
      \textbf{Input:} Pretrained network $\Theta$.
        \vspace{0.1cm}

    \hrule
\begin{algorithmic}[1]
  \STATE {Copy $\Theta$ to obtain student $\Theta_s$ and teacher $\Theta_t$.}
  \STATE {Store BatchNorm statistics $\mu_{\bn{i}}$, $\sigma_{\bn{i}}$ for each BatchNorm layer $\bn{i}$ of $\Theta_s$. Retain them for future use in the data generation function $\mathcal{G}_{\conv{i}}$ (Section~\ref{generation}).}
  \STATE {Perform BatchNorm fusion (Section~\ref{AFCLE}) on $\Theta_s$ and $\Theta_t$.}
  \STATE {Perform Assumption-Free Cross-Layer Equalization (Section~\ref{AFCLE}) on $\Theta_s$ and $\Theta_t$.}
  \STATE {Perform layer-wise compression on $\Theta_s$ using the data generated by $\mathcal{G}_{\conv{i}}$ (see Section~\ref{section:quantization} for quantization, Section~\ref{section:pruning} for pruning).}
  \STATE {Return the compressed model $\Theta_s$}
\end{algorithmic}
\caption{Our layer-wise training algorithm for compressing each layer of a pretrained network $\Theta$ to produce a compressed network.}
\label{alg} 
\end{figure}

\section{Layer-Wise Data-Free Compression}
\begin{figure}[t!]
  \centering
  \includegraphics[width=0.45\textwidth]{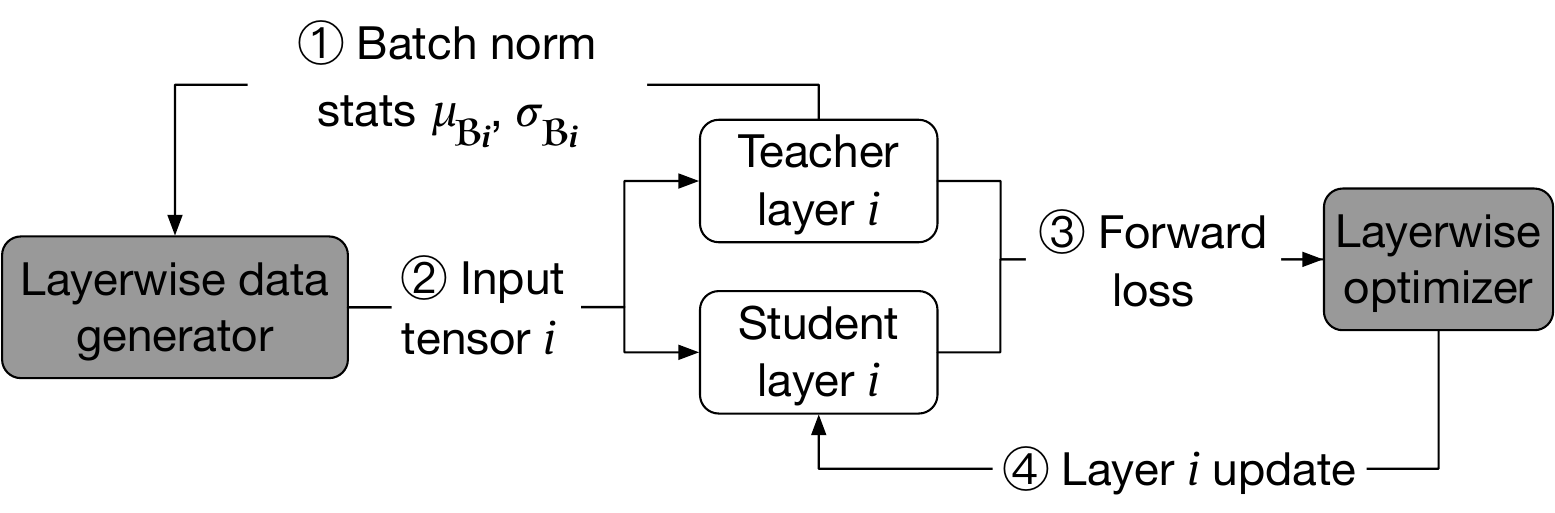}
  \caption{Illustration of optimization of one layer. Note that BatchNorm fusion and AFCLE (not shown) occur as a preprocessing step (Section~\ref{AFCLE}). Our layer-wise optimization is applied separately to each layer of the model.}
  \label{fig:layerwise_optim}
\end{figure}

\begin{figure*}[!t]
\centering
\subfloat[]{ 
  \includegraphics[width=0.3\textwidth]{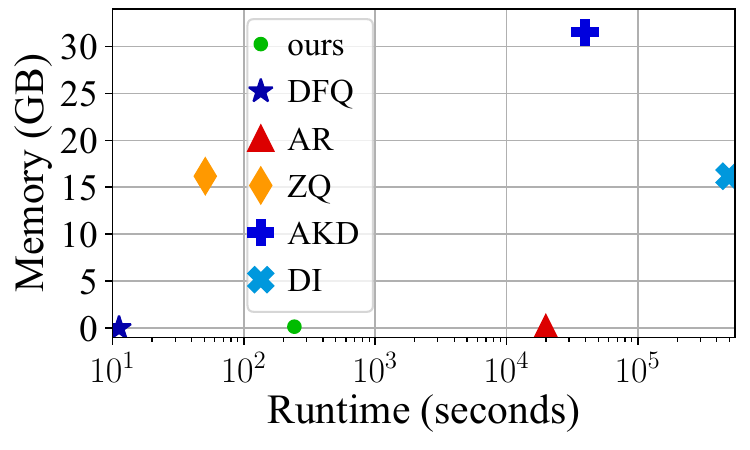}
  \label{fig_first_case}
}
\subfloat[]{
  \includegraphics[width=0.3\textwidth]{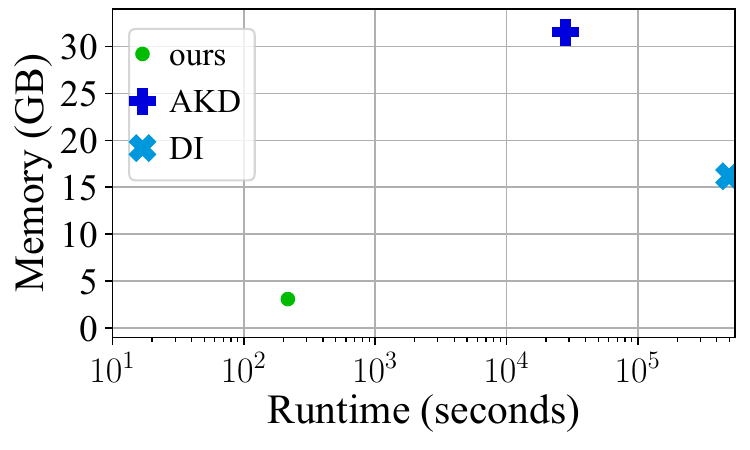}
  \label{fig_second_case}
}
\subfloat[]{    
  \includegraphics[width=0.3\textwidth]{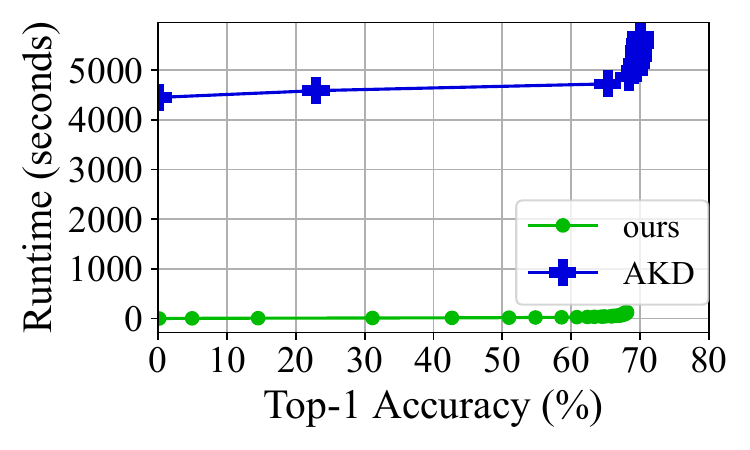}
  \label{fig_third_case}
}

\caption{Runtime and memory overhead for data-free quantization and pruning of MobileNetV1 on an NVidia Tesla V100. (a) Quantization to 8 bits. Note DFQ \cite{DFQ}, AR \cite{AdaRound}, and ZQ \cite{zeroq} are quantization-only methods. (b) Pruning. (c) ImageNet validation accuracy as a function of training time when pruning. DI \cite{DI} requires orders of magnitude more computation than AKD \cite{AKD} (data generation alone takes 492,000 seconds) so we omit it for clarity.} 
\label{fig:runtime_comparison}
\end{figure*}

Our method for data-free network compression begins with a fully trained network and creates a compressed network of the same architecture. This is conceptually similar to Knowledge Distillation \cite{KnowledgeDistillation} in that a pretrained ``teacher'' network is used to train a ``student'' network. However, Knowledge Distillation requires training data. Previous approaches have addressed this through generating data \cite{AKD,DI}, but these methods are computationally expensive (Figure~\ref{fig:runtime_comparison}).

We take a simpler approach illustrated in Figure~\ref{alg} and Figure~\ref{fig:layerwise_optim}. We view each layer of the student as a compressed approximation of the corresponding layer in the teacher. As long as the approximation in each layer is accurate, the overall student network will produce the same outputs as the teacher. Because our data generation doesn't require generating realistic images (unlike AKD \cite{AKD} and DI \cite{DI}), our method takes less time and memory than these approaches (Figure~\ref{fig:runtime_comparison}). Because our method trains layers separately, it doesn't require large gradient buffers needed for backpropagating through the whole neural network, making it more memory efficient than ZeroQ \cite{zeroq}, AKD, and DI. This allows our method to run efficiently on edge devices with limited memory, supporting real-time post-deployment compression. Our method also achieves higher accuracy than baseline low-memory methods AR \cite{AdaRound} and DFQ \cite{DFQ} (Section~\ref{section:experiments}).

Our method is described in detail in the following subsections. We begin by performing fusion and Assumption-Free Cross-Layer Equalization (Section~\ref{AFCLE}). Then, we generate data to use for compression  (Section~\ref{generation}). We then discuss our layer-wise compression algorithms for quantization (Section~\ref{section:quantization}) and pruning (Section~\ref{section:pruning}).

\subsection{Fusion and Equalization} \label{AFCLE}
Two issues complicate the matter of assembling a compressed network from compressed individual layers. For simplicity, we describe these issues in the case of a fully connected layer, but the extension to convolutions is straightforward.

The first issue relates to the BatchNorm \cite{BatchNorm} layers commonly used in CNNs. A BatchNorm layer consists of the parameters $\mu$, $\sigma$, $\gamma$, and $\beta$, which correspond to the mean of its inputs, the standard deviation of its inputs, the weight of its affine transformation, and the bias of its affine transformation. Consider a linear layer $W$ with bias $b$, which is followed by a BatchNorm layer. The output of the linear layer, followed by the BatchNorm layer, is
\begin{equation}
    f(x) = \dfrac{Wx + b - \mu}{\sqrt{\sigma ^2 + \epsilon}} \odot \gamma + \beta,
\end{equation}
where $\epsilon$ is a small number used to avoid division by $0$, and $\odot$ denotes elementwise multiplication \cite{BatchNorm,PyTorch}.

Note that the function $f(x)$ is overparameterized. There are multiple different sets of parameters that yield identical functions $f(x)$. For example, consider the effect of multiplying the $c^{th}$ row of $W$, $b$, and $\mu$ by some nonzero scalar $a$, and multiplying the $c^{th}$ row of $\gamma$ by $1/a$. The adjustments to $W$, $b$, $\mu$, and $\gamma$ will cancel out, and $f(x)$ will remain unchanged.

Thus, the magnitude of rows of $W$ can be rescaled if the corresponding BatchNorm elements are rescaled. This is problematic when pruning or quantizing $W$ because we expect the weight values' magnitudes to reflect their importance \cite{DNW,STR}. This coupling of the values of $W$ with BatchNorm parameters prevents this. To address this issue, we fuse the BatchNorm parameters $\mu$, $\sigma$, $\gamma$, and $\beta$ into the preceding convolution \cite{fusion}, so that the BatchNorm parameters' effective influence on weight magnitudes is accounted for.

The second complication with breaking data-free network compression into layer-wise compression subproblems is that the relative magnitude of weights are not calibrated within a layer. Consider the output of a pair of layers with weights $W_1$ and $W_2$, and biases $b_1$ and $b_2$. Suppose the network uses ReLU activations \cite{ReLU}, such that the output of the pair of layers is
\begin{equation}
    f(x) = \text{ReLU}(W_1 (\text{ReLU} (W_2 x + b_2)) + b_1).
\end{equation}
This function $f(x)$ is overparameterized. If the $c^{th}$ row of $W_2$ is multiplied by a scale factor $a$, and if the $c^{th}$ element of $b_2$ is multiplied by $a$, and if the $c^{th}$ column of $W_1$ is multiplied by $1/a$, then the function $f(x)$ remains unchanged. In this rescaled network, the $c^{th}$ row of $W_2$ has changed, but no other rows in $W_2$ have changed. Thus, the weights' magnitudes do not necessarily reflect their importance to the network, which is problematic for pruning and quantization \cite{DNW,STR}.

To address this scaling inconsistency we employ a method we call Assumption-Free Cross-Layer Equalization (AFCLE). Our method is inspired by Cross-Layer Equalization (CLE) \cite{DFQ}. With each layer $\mathcal{L}_j$ with weight $W_j \in \mathbb{R}^{c_o \times c_i}$ and bias $b_j \in \mathbb{R} ^{c_o}$, we associate a pair of vectors, $v_j^{i} \in \mathbb{R}^{c_i}$ and $v_j^{o} \in \mathbb{R}^{c_o}$. We now compute our layer's output as 
\begin{equation}
    \mathcal{L}_j(x) = (W_j (x \odot v_j^{i}) + b_j) \odot v_j^{o}.
\end{equation}
Each element of the vectors $v_j^{i}$ and $v_j^{o}$ is initialized to $1$. 

AFCLE progresses iteratively across the network. In each iteration, we consider a channel in a pair of adjacent network layers. Consider the $c^{th}$ row of $W_2$ with weights $W_2^c$, and the corresponding $c^{th}$ column of $W_1$ with weights $W_1^c$. Let $[\cdot]_c$ denote the $c^{th}$ element of a vector. We rescale the network as
\begin{align}
    s_c &= \dfrac{\sqrt{\max(|W_1^c|)\max(|W_2^c|)}}{\max(|W_2^c|)} \label{eq:sc_first} \\
    W_1^c &\leftarrow W_1 ^c / s_c \\
    [v_1^{o}]_c &\leftarrow [v_1^{o}]_c * s_c \label{eq:6} \\
    [b_1]_c &\leftarrow [b_1]_c / s_c \label{eq:7} \\
    W_2^c &\leftarrow W_2^c * s_c \\
    [v_2^{i}]_c &\leftarrow [v_2^{i}]_c / s_c. \label{eq:sc_last}
\end{align}

We iterate over all channels and over all pairs of adjacent layers in the network. We continue until the mean of all the scale parameters $s_c$ for one round of equalization deviates from $1$ by less than $10^{-3}$, since weight updates are negligible. 

The choice of $s_c$ is motivated by discussion of CLE in \cite{DFQ}. To summarize, this choice allows the dynamic range of each individual channel to match the dynamic range of the weight tensor to which it belongs, which minimizes the loss of information during compression. See \cite{DFQ} for details.

Our AFCLE method introduces $v_j^{i}$ and $v_j^{o}$, which act as a buffer that reverses the changes to $W_j$ and $b_j$, so that we can equalize $W_j$ across layers without changing the network's outputs at any layer. This allows our method to be used for activation functions that aren't piecewise linear. The main difference between AFCLE and CLE is that our AFCLE method uses $v_j^{i}$ and $v_j^{o}$ to record weight updates, whereas CLE does not (CLE omits Equations \ref{eq:6}, \ref{eq:7}, \ref{eq:sc_last}). CLE assumes that the updates to $W_1^c$ and $W_2^c$ do not alter the network. CLE's assumption holds if the network uses piecewise linear activations.

Note that AFCLE results in no real increase in parameter count, since the vectors $v_j^{i}$ and $v_j^{o}$ can be folded into $W_j$ and $b_j$ after data-free compression is completed. Note also that, when experimenting with networks with only ReLU activations, we simply drop the $v_j^i$ and $v_j^o$ buffers, since they are not needed.

\subsection{Layer-Wise Data Generation} \label{generation}
We now describe our method for generating data used by our layer-wise optimization algorithms. The details of how this generated data is used for quantization and pruning are described in Sections~\ref{section:quantization} and \ref{section:pruning}.

Consider a network composed of blocks containing a convolution, a BatchNorm \cite{BatchNorm}, and an activation. Let $\bn{i}$ denote the BatchNorm layer associated with a block of index $i$. It has stored parameters $\mu_{\bn{i}}$, $\sigma_{\bn{i}}$, $\gamma_{\bn{i}}$, and $\beta_{\bn{i}}$ corresponding to the input mean, input standard deviation, scale factor, and bias. Because the BatchNorm layer normalizes its inputs before applying an affine transformation, the mean of its outputs is $\beta_{\bn{i}}$ and the standard deviation of its outputs is $\gamma_{\bn{i}}$.

We exploit this information to generate layer-wise inputs. Let $\conv{i}$ denote the convolutional layer in block $i$ of the network, and $f_i$ denote the activation function.

Consider the case in which block $i$ accepts multiple input tensors from blocks indexed by elements $j \in \mathcal{K}$. Let $x_{\bn{j}}$ denote the input into the BatchNorm $\bn{j}$ from a training batch (when training with real data). Assuming the inputs to block $i$ are combined by an addition function, the input $x_{\conv{i}}$ to convolution $\conv{i}$ is

\begin{equation}
    x_{\conv{i}} = \sum_{j \in \mathcal{K}} f_{j}(\bn{j}(x_{\bn{j}})).
\end{equation}

During data-free compression, we do not have access to $x_{\bn{j}}$, so we estimate it. Let $\mathcal{G}_{\conv{i}}$ be a function that generates an input used to train layer $\conv{i}$. Using our observation above regarding the output statistics of BatchNorms, we estimate

\begin{align}
    x_{\bn{j}} &\sim \mathcal{N}(\beta_{\bn{j}}, \gamma_{\bn{j}}) \\
    \mathcal{G}_{\conv{i}} &= \sum_{j \in \mathcal{K}} f_j (x_{\bn{j}}) \label{generation_equation},
\end{align}
where $\mathcal{N}(a, b)$ denotes a Gaussian function with mean $a$ and standard deviation $b$. If a layer is not preceded by a BatchNorm, we generate data from $\mathcal{N}(0, 1)$ (as in the case of the network's first layer, or if BatchNorms aren't present before the convolution). We ignore the effect of other layers (such as Average Pooling). Note that BatchNorm statistics need to be gathered before BatchNorm fusion (Section~\ref{AFCLE}) because the parameters are modified during fusion.

\subsection{Data-Free Quantization} \label{section:quantization}

\begin{figure}[t]
\hrule
\vspace{0.1cm}
    \textbf{Input:} BatchNorm parameters $\beta_{\bn{j}},\gamma_{\bn{j}}$, input activation functions $f_j$, number of steps $N$, quantization bits $b$.
        \vspace{0.1cm}

    \hrule
\begin{algorithmic}[1]
  \STATE {$\mathcal{G}_{\conv{i}} = \sum_{j \in \mathcal{K}} f_j (\mathcal{N}(\beta_{\bn{j}}, \gamma_{\bn{j}}))$ (Equation~\ref{generation_equation})}
  \STATE {$X \sim \mathcal{G}_{\conv{i}}$}
  \STATE {$x_{\max} \gets \max(X)$, $x_{\min} \gets \min(X)$}
  \STATE {$h \gets -\infty$, $l \gets \infty$, $L \gets \infty$}
  \FOR {$h_i \in [1, 2, ..., N]$}
  \FOR {$l_i \in [1, 2, ..., N]$}
    \STATE {$\tilde{h} = (h_i / N) (\max(x_{\max}, 0))$}
    \STATE {$\tilde{l} = (l_i / N) (\min(x_{\min}, 0))$}
  \IF {$||X - Q(X, \tilde{l}, \tilde{h}, b)||_2 < L$}
    \STATE {$l \gets \tilde{l}$, $h \gets \tilde{h}$}
    \STATE {$L \gets ||X - Q(X, \tilde{l}, \tilde{h}, b)||_2$}
  \ENDIF
  \ENDFOR
  \ENDFOR
  \STATE {return $l$, $h$}
\end{algorithmic}
\caption{Our method for data-free quantization. We compute the high end $h$ and the low end $l$ of the activation range for our activation quantizers without using data. This corresponds to the ``Layerwise Optimizer'' step in Figure~\ref{fig:layerwise_optim}.}
\label{alg:quantization_algorithm}
\end{figure}
We now describe how to use our generated data to quantize a neural network. Our method quantizes both weights and activation ranges. As is standard, we set our weight quantizers to the minimum and maximum of the weight tensors \cite{google_quantization}.

We set activation ranges by performing a simple optimization (Figure~\ref{alg:quantization_algorithm}). Recall that the quantized version of a floating-point activation tensor $X$ can be expressed as $Q(X, l, h, b)$:
\begin{align}
    I(X, l, h, b) &\equiv \left\lfloor \dfrac{\min(\max(X, l) ,h) - l}{(h-l)/(2^b-1)} \right\rceil \\
    Q(X, l, h, b) &= \dfrac{h-l}{2^b-1} I(X, l, h, b) + l, \label{eq:quantization}
\end{align}
where $l$ is the minimum of the quantization range, $h$ is the maximum, $b$ is the number of bits in the quantization scheme, and $\lfloor \cdot \rceil$ denotes rounding to the nearest integer. We perform a grid search jointly over $h \in [0, \max(X_{\max}, 0]]$ and $l \in [\min(0, X_{\min}), 0]$ to minimize $|X - Q(X, l, h, b)|_2$.

Our overall method for data-free quantization follows Figure~\ref{alg}. First, we perform BatchNorm fusion and AFCLE on the student network (Section~\ref{AFCLE}). Then, we perform bias absorption, as in \cite{DFQ}. Next, we set activation quantizers (Figure~\ref{alg:quantization_algorithm}). Then, we perform bias correction, as in \cite{DFQ}. We then set activation quantizers again (Figure~\ref{alg:quantization_algorithm}), because the bias correction step adjusted weights and output statistics slightly.

\subsection{Data-Free Pruning} \label{section:pruning}
We now describe how to use our generated data (Section~\ref{generation}) to prune neural network weights. Our overall method follows Figure~\ref{alg}. We begin by duplicating our pretrained network $\Theta$ to obtain a teacher $\Theta_t$ and a student $\Theta_s$. We perform fusion and AFCLE as a preprocessing step (Section~\ref{AFCLE}). For the remainder of our method, the teacher will remain unchanged, and the student will be pruned.

We prune using Soft Threshold Reparameterization (STR) \cite{STR} with gradient descent \cite{backprop}. In convolutional and fully-connected layers, an intermediate weight $W_s$ is calculated as
\begin{equation}
    W_s = \text{sign} (W) \cdot \text{ReLU}(|W| - \text{sigmoid}(s)),
\end{equation}
where $W$ is the weight tensor, and $s$ is a model parameter used to control sparsity. This intermediate tensor $W_s$ is used in place of $W$ in the forward pass. In the backward pass, the gradients are propagated to the original weight matrix $W$. A weight decay parameter $\lambda$ is used to drive $s$ upwards from an initial value $s_0<0$, which increases sparsity \cite{STR}.

\begin{table*}[ht]
\renewcommand{\arraystretch}{1}
\setlength{\tabcolsep}{4pt}
\begin{center}
\caption{
ImageNet results comparing quantization methods. ($\dagger$) denotes relatively memory-intensive methods (see Figure~\ref{fig:runtime_comparison}). Our method achieves accuracies matching or surpassing baselines in most cases. Generative methods (AKD and DI) fail to produce high-accuracy solutions at low bit widths. For MobileNets, our accuracy gains increases at lower bit widths.
\label{fig:quantized_efficient}
}
\begin{tabular}{c | c c c c c c | c c c c c c | c c c c c c }
Bits & ours & DFQ & AR & ZQ$^\dagger$ & AKD$^\dagger$ & DI$^\dagger$ & ours & DFQ & AR & ZQ$^\dagger$ & AKD$^\dagger$ & DI$^\dagger$ & ours & DFQ & AR & ZQ$^\dagger$ & AKD$^\dagger$ & DI$^\dagger$ \\
\hline
& \multicolumn{6}{c|}{MobileNetV1} & \multicolumn{6}{c|}{MobileNetV2} & \multicolumn{6}{c}{ResNet18} \\
\hline
8  & 71.13 & 71.18 & \textbf{71.35} & 64.36 & 70.13 & 24.53 &   \textbf{70.50} & 70.16 & 68.61 & 62.03 & 19.33 & 7.21 & \textbf{68.67} & 67.72 & 56.24 & 43.69 & 62.13 & 19.69\\
7  & \textbf{70.26} & 69.76 & 70.22 & 54.31 & 67.77 & 21.35 &   \textbf{69.67} & 68.89 & 66.65 & 45.42 & 4.87 & 5.43  & \textbf{67.92} & 66.40 & 54.45 & 39.65 & 61.38 & 19.05\\
6  & \textbf{66.95} & 61.81 & 63.68 & 28.23 & 7.11 & 19.17  & \textbf{66.97} & 63.06 & 51.00 & 5.16 & 9.70 & 3.54     & \textbf{63.72} & 62.86 & 50.57 & 41.12 & 54.96 & 16.98\\
5  & \textbf{55.37} & 25.63 & 29.17 & 0.51 & 0.10 & 9.06    & \textbf{56.33} & 23.41 & 27.83 & 1.98 & 0.86 & 2.62     & 35.47 & 34.76 & \textbf{43.64} & 12.74 & 41.41 & 12.43\\
4  & \textbf{5.62} & 0.23 & 0.28 & 0.09 & 0.10 & 0.10       & \textbf{8.23} & 0.49 & 0.48 & 0.07 & 0.10 & 0.09        & 0.88 & 1.06 & \textbf{10.64} & 0.14 & 0.10 & 6.33\\
\end{tabular}
\end{center}
\end{table*}
\begin{figure*}[t!]
\vspace{-.4cm}
\includegraphics[width=.9\textwidth]{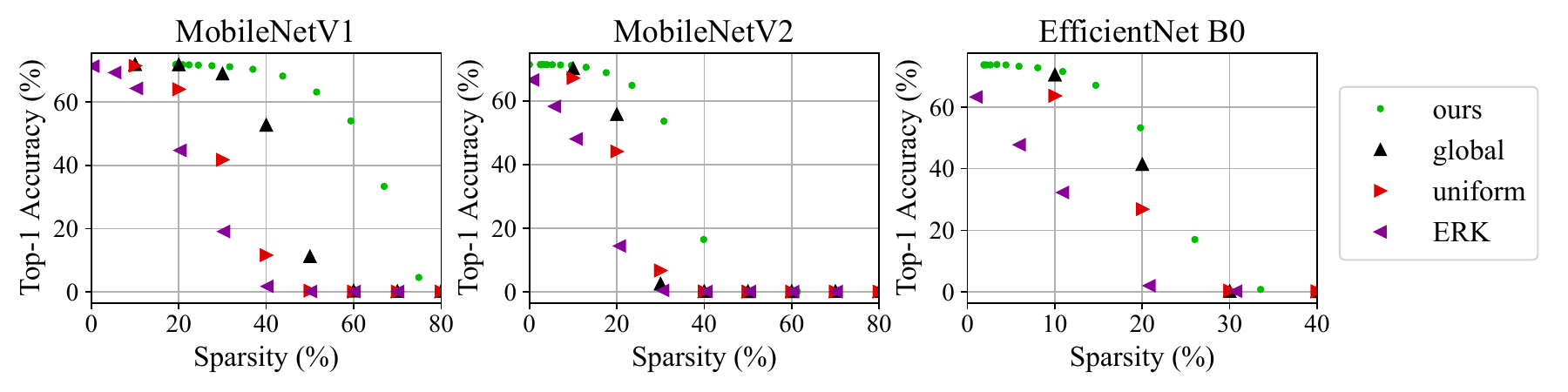}
\vspace{-.35cm}
  \caption{
  ImageNet results comparing efficient pruning methods. Our method produces the strongest sparsity/accuracy tradeoff. For example, in MobileNetV1, we improve the sparsity from $40\%$ (global) to $60\%$ (ours) for models with a Top-1 accuracy of $50\%$.
  }
  \label{efficient_pruning}
\end{figure*}

We prune each layer of $\Theta_s$ separately. Let $i$ be an index assigned to the convolutional or fully-connected layer $\mathcal{C}_i^s$ in the student, and let the corresponding layer in the teacher be $\mathcal{C}_i^t$. Let $\mathcal{C}_i^s(\cdot)$ denote the application of a layer to an input tensor. Given a loss function $L_i$ associated with layer $i$, we compute the loss for layer $i$ as
\begin{equation}
    L = L_i(\mathcal{C}_i^s(x), \mathcal{C}_i^t(x)), \label{layerwise-loss}
\end{equation}
where $x \sim \mathcal{G}_{\mathcal{C}_i^t}$ is generated from Equation~\ref{generation_equation}. In our experiments, we choose $L_i$ to be the mean square error loss for each $i$. We also freeze the student's bias during training, since our goal is to induce sparsity only in the weights. We empirically found that including the activations $f_j$ in Equation~\ref{generation_equation} is not important during pruning, so we omit them. 

\section{Experiments} \label{section:experiments}
We provide results for data-free quantization and pruning. We separately discuss efficient methods and expensive methods to clarify which results are obtainable with on-device compression in the low-compute scenario. We evaluate on ImageNet \cite{ImageNet}. We train in PyTorch \cite{PyTorch} using NVIDIA Tesla V100 GPUs. When training with backpropagation, we use Adam \cite{Adam} with a cosine learning rate decaying from $0.001$ to $0$ over $10^5$ iterations (though in practice, our method converges in only a few hundred iterations). We use batch size $128$ for all methods, reducing it to fit on a single GPU as needed.

In all MobileNetV2 \cite{MobileNetV2} experiments, we replace ReLU6 with ReLU \cite{ReLU} as in DFQ \cite{DFQ}. The accuracy of this modified network is unchanged. For all experiments, we place activation quantizers after skip connections, and we quantize per-tensor (not per-channel) as in Equation~\ref{eq:quantization}. This differs from previous works \cite{zeroq}, so our quantization baselines differ slightly. 

\subsection{Data-Free Quantization} \label{section:dfq}

\textbf{Efficient Quantization:}
In Table~\ref{fig:quantized_efficient}, we compare our results with methods with a similar compute envelope. When generating data for quantizing activations (Figure~\ref{alg:quantization_algorithm}), we use a batch size of 2000, and optimize for $N=100$ steps. We evaluate MobileNetV1 \cite{MobileNet}, MobileNetV2 \cite{MobileNetV2}, and ResNet18 \cite{ResNet}.

We compare to DFQ \cite{DFQ}. DFQ performs preprocessing similar to our method, but simply sets activation quantizers to be $6$ standard deviations from the mean (using BatchNorm statistics). We also compare to AdaRound (AR) \cite{AdaRound}, which optimizes over rounding decisions. AR requires training data, but we use our generated data for a fair comparison.

Note that, since our method and DFQ don't require backpropagation, we simply report the accuracy. For AR, which requires backpropagation, we report the final-epoch accuracy. We do this since a truly data-free scenario would not allow for a validation set on which to perform early stopping. Our method outperforms DFQ and AR in nearly all cases.

\textbf{Expensive Quantization:}
We evaluate more memory-intensive methods ZeroQ (ZQ) \cite{zeroq}, Adversarial Knowledge Distillation (AKD) \cite{AKD}, and Deep Inversion (DI) \cite{DI} in Table~\ref{fig:quantized_efficient}. As before, we report final-epoch accuracies. Our method outperforms these methods using orders of magnitude less memory (Figure~\ref{fig:runtime_comparison}) in almost all cases. DI failed to produce accuracies above $25\%$. AKD failed to produce good results on MobileNets at $6$ bits and below, whereas our method produced strong results across all networks. Note that AKD requires training a GAN, which may require parameter tuning for different network architectures or datasets. Such tuning is not possible in a truly data-free setup. 

\subsection{Data-Free Pruning}
\begin{figure*}[ht!]
  \includegraphics[width=.9\textwidth]{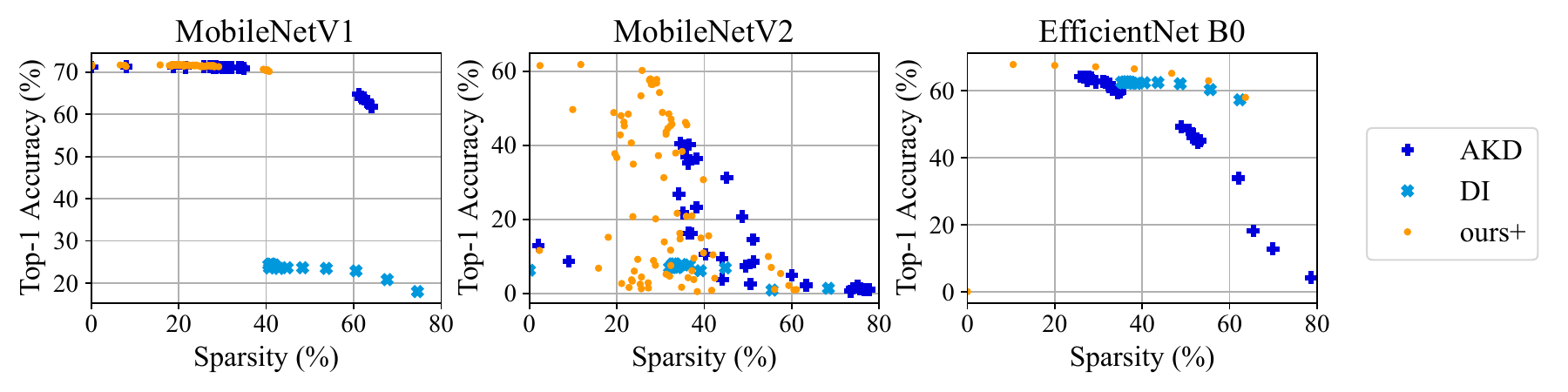}
  \vspace{-.35cm}
  \caption{Comparison of computationally expensive pruning methods on ImageNet. DI fails to converge to stable solutions on MobileNets. On EfficientNet B0, combining our method with DI achieves stronger results than DI alone.
  }
  \label{expensive_pruning}
\end{figure*}

\begin{figure}[t!]
  \centering
  \includegraphics[width=0.4\textwidth]{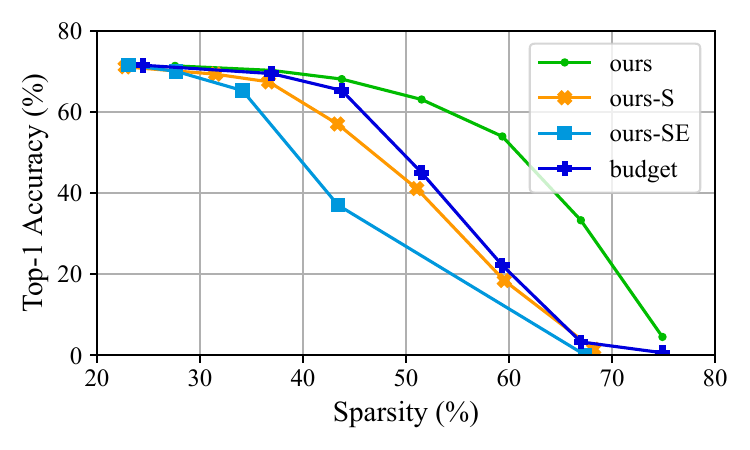}
  \vspace{-.35cm}
  \caption{
Ablation study of pruning MobileNetV1 on ImageNet. Removing input scaling (ours-S) reduces performance. Further removing equalization (ours-SE) incurs additional loss of accuracy. ``Budget'' refers to using our learned sparsity pattern, but with the original pretrained network weights.
  }
  \label{fig:mnv1_ablation}
\end{figure}
\textbf{Efficient Pruning:} We present results for our pruning method in Figure~\ref{efficient_pruning}. We fix the sparsity-inducing weight decay parameter to $\lambda=1.55 \times 10^{-5}$, as in \cite{STR}. We investigate MobileNets as before. We also investigate EfficientNet \cite{EfficientNet} to include results for networks with non-ReLU activations. We did not investigate EfficientNet for quantization because several baseline methods required only ReLU activations.

Because our efficient quantization baselines do not support pruning, we compare to different baselines. ``Global'' corresponds to pruning every weight whose magnitude is smaller than a given threshold \cite{WITS}. ``Uniform'' corresponds to applying a uniform sparsity level to each layer by pruning the weights with smallest magnitude \cite{WITS}. The Erdosh-Renyi Kernel (ERK) baseline corresponds to a budgeted layer-wise pruning \cite{RIGL}. Each layer's fraction of pruned weights is
\begin{equation} \label{eq:12}
    p = \dfrac{c_{o} + c_i + k_h + k_w}{c_o c_i k_h k_w},
\end{equation}
where $c_o, c_i$ are the number of output and input dimensions, and $k_h, k_w$ are the kernel height and width. For these baselines, we perform BatchNorm fusion \cite{fusion}  because it is a standard technique, but we do not perform AFCLE.

Our method outperforms these baselines, producing a better sparsity/accuracy tradeoff. In Figure~\ref{fig:mnv1_ablation}, we present an ablation. ``Ours-S'' represents generating data from $\mathcal{N}(0, 1)$, with no other changes to our method. ``Ours-SE'' represents generating data from $\mathcal{N}(0, 1)$ and skipping AFCLE. ``Budget'' represents a model that uses the layer-wise pruning budget learned by our method (without retraining weights). We find that generating data from $\mathcal{N}(0, 1)$ and skipping AFCLE both reduce performance substantially.

\textbf{Expensive Pruning:}
We compare our method to more computationally expensive methods in Figure~\ref{expensive_pruning}. Our baselines are AKD \cite{AKD} and DI \cite{DI} (note that ZeroQ \cite{zeroq} does not support pruning). These methods involve generating end-to-end training data. We found we needed to sweep across more values of the sparsity-controlling weight decay parameter $\lambda$ to encourage varying levels of sparsity in AKD. In addition to $\lambda=1.55*10^{-5}$, we used $\{2 \lambda, 10 \lambda, 100 \lambda\}$. We report final-epoch accuracies. For each network, we combined our method with the top-performing baseline method for that network (``ours+'', Figure~\ref{expensive_pruning}). To do this, we add our layer-wise loss (Equation~\ref{layerwise-loss}) to the baseline's loss function and backpropagate as usual (we omit fusion and AFCLE for simplicity in this case).

Because STR controls the sparsity level implicitly (not explicitly), it does not give direct control over the final sparsity levels. This is why some methods never converge to lower-sparsity, higher-accuracy solutions (for example, DI for MobileNetV1). For MobileNets, DI failed to produce solutions with more than $30\%$ accuracy. For AKD, performance was stronger in general, although solutions for MobileNetV2 did not exceed $40\%$ accuracy. Some parameter settings for AKD for MobileNetV2 resulted in very poor performance.

For MobileNetV1, we combine our method with AKD (``ours+'', Figure~\ref{expensive_pruning}), producing a slight increase in accuracy for moderately sparse models. For MobileNetV2, we combine our method with AKD to produce higher-accuracy solutions. For EfficientNet B0, we combine our method with DI and improve the overall sparsity/accuracy tradeoff.

These methods achieve a stronger sparsity/accuracy tradeoff than efficient pruning methods, but getting these methods to work may be difficult in practice without data. AKD and DI use many more parameters than the efficient methods. These parameters may need more tuning to work on data from another domain or for other architectures. Moreover, many trials of AKD and DI converged to low accuracy, which isn't detectable without data.

\section{Conclusion}
We present an efficient, effective method for data-free compression. We break this problem into the subproblem of compressing individual layers without data. We precondition networks to better maintain accuracy during compression. Then, we compress individual layers using data generated from BatchNorm \cite{BatchNorm} statistics from previous layers. Our method outperforms baselines on quantization. Our method outperforms other computationally efficient methods for pruning, and can be combined with computationally expensive methods to improve their results.

\IEEEtriggeratref{48}


\bibliographystyle{IEEEtran}
\bibliography{egbib}

\end{document}